\documentclass{article}

\usepackage[final]{nips_2018}

\usepackage[utf8]{inputenc} 
\usepackage[T1]{fontenc}    

\usepackage{amsmath}
\usepackage{hyperref}       
\usepackage{url}            
\usepackage{booktabs}       
\usepackage{amsfonts}       
\usepackage{nicefrac}       
\usepackage{microtype}      
\usepackage{graphicx} 
\usepackage{amsthm}
\usepackage{amssymb}
\usepackage{array}
\usepackage{multirow}
\usepackage{subcaption}
\usepackage{float}
\usepackage[mathscr]{euscript}
\usepackage{algorithm}
\usepackage{algpseudocode}
\usepackage{xcolor}
\usepackage{color}
\newcommand{\vertiii}[1]{{\left\vert\kern-0.25ex\left\vert\kern-0.25ex\left\vert #1 
		\right\vert\kern-0.25ex\right\vert\kern-0.25ex\right\vert}}
\newcommand{\be}[1]{\begin{equation}\label{#1}}
\newcommand{\benon}{\begin{equation*}}  
\newcommand{\bemuln}[1]{\begin{multline}\label{#1}}
\newcommand{\bemul}{\begin{multline*}}
\newcommand{\bee}{\begin{eqnarray*}}
\newcommand{\eee}{\end{eqnarray*}}
\newcommand{\been}[1]{\begin{eqnarray}\label{#1}}
\newcommand{\eeen}{\end{eqnarray}}
\newcommand{\began}[1]{\begin{gather}\label{#1}}
\newcommand{\bega}{\begin{gather*}}
\newcommand{\bealn}[1]{\begin{align}\label{#1}}
\newcommand{\beal}{\begin{align*}}
\newcommand{\bealatn}[2]{\begin{alignat}{#1}\label{#2}}
\newcommand{\bealat}{\begin{alignat*}}
\newcommand{\bexalatn}[1]{\begin{xalignat}\label{#1}}
\newcommand{\bexalat}{\begin{xalignat*}}

\newcommand{\mbb}{\mathbb}

\newtheorem{thm}{Theorem}[section]

\newtheorem{defi}{Definition}

\def\bx{{\mathbf x}}  
\def\by{{\mathbf y}}

\def\bW{{\mathbf W}}

\def\texitem#1{\par\smallskip\noindent\hangindent 25pt
               \hbox to 25pt {\hss #1 ~}\ignorespaces}

\newcommand{\bbeta}{\boldsymbol{\beta}}

\title{Learning Optimal Personalized Treatment Rules Using Robust Regression Informed K-NN}

\author{
  Ruidi Chen\\
  Division of Systems Engineering, \\
  	Boston University, Boston, MA \\
  	\texttt{rchen15@bu.edu}.
  \And 
  Ioannis Ch. Paschalidis \\
  Department of Electrical and Computer Engineering, \\
  	Division of Systems Engineering, \\
  	and Department of Biomedical Engineering, \\
  	Boston University, Boston, MA \\
  	\texttt{yannisp@bu.edu}
 }

\begin{document}

\maketitle

\begin{abstract}
  We develop a prediction-based prescriptive model for learning optimal personalized treatments for patients based on their Electronic Health Records (EHRs). Our approach consists of: (i) predicting future outcomes under each possible therapy using a robustified nonlinear model, and (ii) adopting a randomized prescriptive policy determined by the predicted outcomes. We show theoretical results that guarantee the out-of-sample predictive power of the model, and prove the optimality of the randomized strategy in terms of the expected true future outcome. We apply the proposed methodology to develop optimal therapies for patients with type 2 diabetes or hypertension using EHRs from a major safety-net hospital in New England, and show that our algorithm leads to a larger reduction of the HbA\textsubscript{1c}, for diabetics, or systolic blood pressure, for patients with hypertension, compared to the alternatives. We demonstrate that our approach outperforms the standard of care under the robustified nonlinear predictive model.
\end{abstract}

\section{Introduction}
Learning the optimal treatment policies from large amounts of clinical data presents challenges due to (i) the significant ``noise'' caused by recording errors, missing values, and large variability across patients, (ii) the lack of counterfactual information, and (iii) the underlying nonlinearity whose parametric form is not known a priori. Attempts on handling these issues have been focusing on denoising \cite{atan2018deep, shi2016robust} and robust nonlinear learning \cite{zhao2014doubly, shi2018high}. A popular research direction adopts a dynamic point of view and models the optimal treatment problem through reinforcement learning \cite{chakraborty2010inference, shortreed2011informing, prasad2017reinforcement, raghu2017continuous} and A-learning \cite{murphy2003optimal, shi2018high}. There are also works adopting a Bayesian approach \cite{lee2015bayesian, wang2016optimal}, deep learning \cite{atan2018deep, liang2018deep}, and miscellaneous statistical inference/machine learning methods \cite{wang2018quantile, shi2018maximin, wang2016learning, wang2018learning, he2018drug} for constructing effective prescriptive rules.

We are interested in learning optimal treatment decisions from Electronic Health Records (EHRs) using computationally efficient and interpretable predictive algorithms that address the aforementioned challenges. We start with a regularized {\em Least Absolute Deviation (LAD)} \cite{wang2006regularized} model that is obtained as a result of a {\em Distributionally Robust Optimization (DRO)} \cite{chen2017outlier} problem and thus hedges against significant noise and the presence of outliers. We then predict future outcomes under each treatment by identifying neighbors through an LAD-weighted metric and fitting a {\em K-Nearest Neighbors (K-NN)} regression model \cite{altman1992introduction} that captures the local nonlinearity in a non-parametric way. A randomized policy is then developed to prescribe each therapy with a probability inversely proportional to its exponentiated predicted outcome.

Our method is similar to \cite{bertsimas2017personalized} where the K-NN regression with an {\em Ordinary Least Squares (OLS)}-weighted metric was used to learn the optimal treatment for type 2 diabetic patients. The key differences lie in that: (i) we adopt a robustified regression procedure that is immunized against high noise and is thus more stable and reliable; (ii) we propose a randomized prescriptive policy that adds robustness to the methodology whereas \cite{bertsimas2017personalized} deterministically prescribed the treatment that was predicted to result in the lowest HbA\textsubscript{1c}; (iii) we show rigorous theoretical results that guarantee the optimality of the randomized strategy, and (iv) the prescriptive rule in \cite{bertsimas2017personalized} was activated when the improvement of the recommended treatment over the standard of care exceeded a certain threshold whereas our method looks into the improvement over the previous regimen. This distinction makes our algorithm applicable in the scenario where the standard of care is unknown or ambiguous. Further, we derive a closed-form expression for the threshold level, which greatly improves the computational efficiency compared to \cite{bertsimas2017personalized} where a threshold was selected by cross-validation.

\section{Methods}
Given a training set of EHRs, we group the patients based on the treatment they receive at the time of the visit. Within each treatment group $m \in [M]$, where $[M] \triangleq \{1, \ldots, M\}$, denote by $\bx_m \in \mbb{R}^p$ a patient-specific feature vector that includes the patient characteristics (e.g., gender, age, race, and {\em Body Mass Index (BMI)}) and diagnosis history. Our goal is to predict the future outcome $y_m$ under treatment $m$. Assume the following nonlinear regression model:
\begin{equation*} 
y_m = \bx_m'\bbeta^*_m + h_m(\bx_m) + \epsilon_m,
\end{equation*}
where prime denotes transpose, $\bbeta^*_m$ is the coefficient vector that captures the linear trend, $h_m(\bx_m)$ describes the nonlinear fluctuation in $y_m$, and $\epsilon_m$ is the noise term with zero mean and standard deviation $\eta_m$ that expresses the intrinsic randomness of $y_m$ and is assumed to be independent of $\bx_m$. 

We follow the robust predictive procedure proposed in \cite{chen2017outlier} to estimate the linear coefficient $\bbeta^*_m$. It reduces to solving the following regularized LAD problem:
\begin{equation} \label{qcp}
\inf\limits_{\bbeta_m} \frac{1}{N_m}\sum\limits_{i=1}^{N_m}|y_{mi} - \bx_{mi}'\bbeta_m| + r_m\|(-\bbeta_m, 1)\|_2,
\end{equation}
where $(\bx_{mi}, y_{mi}), i \in [N_m]$, are the feature-outcome vectors of patients who receive treatment $m$, and $r_m$ is a regularization penalty. 
Solving (\ref{qcp}) gives us a robust estimator of $\bbeta_m^*$, which we denote by $\hat{\bbeta}_m$. It measures the relative significance of the features in terms of their relevance to $y_m$, and will be used to identify the nearest neighbors in the nonlinear non-parametric K-NN regression model. Specifically, we consider the following weighted distance metric,
\begin{equation} \label{wtd}
\|\bx - \bx_{mi}\|_{\hat{\bW}_m}^2 = (\bx-\bx_{mi})'\hat{\bW}_m (\bx-\bx_{mi}),
\end{equation} 
where $\hat{\bW}_m$ is a diagonal matrix with diagonal elements $(\hat{\beta}_{m1})^2, \ldots, (\hat{\beta}_{mp})^2$, where $\hat{\beta}_{mi}$ is the $i$-th element of $\hat{\bbeta}_m$. Given a patient with features $\bx$, we find her $K_m$ nearest neighbors using (\ref{wtd}) within each treatment group $m$. The predicted future outcome for $\bx$ if treatment $m$ is prescribed, denoted by $\hat{y}_m(\bx)$, is computed as the average outcome among the $K_m$ nearest neighbors, i.e.,
\begin{equation} \label{knn}
\hat{y}_m(\bx) = \frac{1}{K_m}\sum_{i=1}^{K_m} y_{m(i)},
\end{equation}  
where $y_{m(i)}$ is the outcome of the $i$-th nearest neighbor to $\bx$.  
(\ref{knn}) can be viewed as a local smooth estimator in the neighborhood of $\bx$. The selected neighbors are similar to $\bx$ in the features that are predictive of the outcome.

Next consider a randomized policy that prescribes treatment $m$ with probability $e^{-\xi \hat{y}_m(\bx)}/\sum_{j=1}^M e^{-\xi \hat{y}_j(\bx)}$, with $\xi$ some pre-specified positive scalar. Theorem \ref{thm:random} characterizes the expected {\em true} outcome under this policy.

\begin{thm} \label{thm:random}
	Given any fixed predictor $\bx \in \mbb{R}^p$, denote its predicted and true future outcome under treatment $m$ by $\hat{y}_m(\bx)$ and $y_m(\bx)$, respectively. Assume that we adopt a randomized strategy that prescribes treatment $m$ with probability $e^{-\xi \hat{y}_m(\bx)}/\sum_{j=1}^M e^{-\xi \hat{y}_j(\bx)},$ for some $\xi \ge 0$. Assume $\hat{y}_m(\bx)$ and $y_m(\bx)$ are non-negative, $\forall m \in [M]$. The expected true outcome under this policy satisfies:
	\begin{equation} \label{eq:random}
	\begin{aligned}
	\sum_{m=1}^M \frac{e^{-\xi \hat{y}_m(\bx)}}{\sum_j e^{-\xi \hat{y}_j(\bx)}}  y_m(\bx) & \le y_k(\bx) + \bigg(\hat{y}_k(\bx)  - \frac{1}{M} \sum_{m=1}^M \hat{y}_m(\bx) \bigg) \\
	& + \xi \bigg( \frac{1}{M} \sum_{m=1}^M \hat{y}_m^2(\bx) + \sum_{m=1}^M \frac{e^{-\xi \hat{y}_m(\bx)}}{\sum_j e^{-\xi \hat{y}_j(\bx)}}  y_m^2(\bx)\bigg) + \frac{\log M}{\xi}, 
	\end{aligned}
	\end{equation}
	for any $k \in [M]$.
\end{thm}

Theorem \ref{thm:random} says that the expected {\em true} outcome of the randomized policy is no worse than the true outcome of any treatment $k$ plus two components, one accounting for the gap between the {\em predicted} outcome under $k$ and the average predicted outcome, and the other depending on the parameter $\xi$. Thinking about choosing $k = \arg \min_m y_m(\bx)$, if $\hat{y}_k(\bx)$ is below the average predicted outcome (which should be true if we have an accurate prediction), it follows from (\ref{eq:random}) that the randomized policy leads to a nearly optimal future outcome by an appropriate choice of $\xi$. 

In consideration of the health care costs and treatment transients, it is not desired to switch patients' treatments too frequently. We thus set a threshold level for the expected improvement on the outcome, below which the randomized strategy will be frozen and the current therapy will be continued. Specifically, 
\begin{equation*} 
m_{\text{f}}(\bx) = 
\begin{cases}
\text{$m$, w.p. $\frac{e^{-\xi \hat{y}_m(\bx)}}{\sum_{j=1}^M e^{-\xi \hat{y}_j(\bx)}}$}, \ \text{if $\sum_{k} \frac{e^{-\xi \hat{y}_k(\bx)}}{\sum_j e^{-\xi \hat{y}_j(\bx)}}  \hat{y}_k(\bx) \le x_{\text{co}} - T(\bx)$}, \\
m_{\text{c}}(\bx), \qquad \qquad \qquad \  \text{otherwise},
\end{cases}
\end{equation*}
where $m_{\text{f}}(\bx)$ and $m_{\text{c}}(\bx)$ are the future and current treatments of patient $\bx$, respectively; $x_{\text{co}}$ represents the current observed outcome, which is assumed to be one of the components of $\bx$, and $T(\bx)$ is some threshold level which is determined in Theorem \ref{threshold}. 

\begin{thm} \label{threshold}
	Assume that the distribution of the predicted outcome $\hat{y}_m(\bx)$ conditional on $\bx$, is sub-Gaussian, and its $\psi_2$-norm is equal to $\sqrt{2}C_m(\bx)$, for any $m \in [M]$ and any $\bx$. Given a small $0 < \bar{\epsilon} < 1$, if
	\begin{equation*}
	\mbb{P}\Bigl(\sum_k \frac{e^{-\xi \hat{y}_k(\bx)}}{\sum_j e^{-\xi \hat{y}_j(\bx)}}  \hat{y}_k(\bx) > x_{\text{co}} - T(\bx)\Bigr) \le \bar{\epsilon},
	\end{equation*}
	then,
	\begin{equation*}
	T(\bx) = \max\Bigl(0, \ \min_m \Bigl(x_{\text{co}} - \mu_{\hat{y}_{m}}(\bx) - \sqrt{-2 C_m^2(\bx)\log (\bar{\epsilon}/M)}\Bigr)\Bigr),
	\end{equation*}
	where $\mu_{\hat{y}_m}(\bx) = \mbb{E}[\hat{y}_m(\bx)|\bx]$.
\end{thm}

Notice that as $\xi \to \infty$, the randomized policy will assign probability $1$ to the treatment with the lowest predicted outcome, which is equivalent to the deterministic policy used in \cite{bertsimas2017personalized}. In this case, slight modification to the threshold level $T(\bx)$ is given as follows:
\begin{equation*}
T(\bx) = \max\Bigl(0, \ \min_m \Bigl(x_{\text{co}} - \mu_{\hat{y}_{m}}(\bx) - \sqrt{-2 C_m^2(\bx)\log \bar{\epsilon}}\Bigr)\Bigr).
\end{equation*}

\section{Results}
We apply our method to develop optimal prescriptions for patients with type-2 diabetes and/or hypertension. 
The data used for the study come from the Boston Medical Center -- the largest safety-net hospital in New England -- and consist of EHRs containing the patients' medical history in the period 1999--2014. We consider the following sets of features for building the predictive model: (i) demographic information, including gender, age and race; (ii) measurements, including systolic blood pressure and diastolic blood pressure, Body Mass Index (BMI) and heart rate; (iii) lab tests, including blood chemistry tests and hematology tests, and (iv) diagnosis history. For the diabetic patients, we consider both oral, e.g., metformin, pioglitazone, and sitagliptin, etc., and injectable prescriptions, e.g., insulin. A complete list of the oral prescriptions can be found in the Appendix. For the hypertension patients, six types of prescriptions are considered: ACE inhibitor, Angiotensin Receptor Blockers (ARB), calcium channel blockers, thiazide and thiazide-like diuretics, $\alpha$-blockers and $\beta$-blockers.  

The diabetes dataset contains 12,016 patient visits with 63 normalized features, and the hypertension dataset contains 26,128 patient visits with the same set of features. We randomly split both datasets into a training (80\%) and a test set (20\%), where the training set is used to tune the parameters (e.g., the $\ell_2$ regularizer $r_m$, the number of neighbors $K_m$, and the exponent $\xi$ in the randomized policy) and train the models, and the test set is used to evaluate the performance. We regress the tuned $K_m$ against $\sqrt{N_m}$ and use the derived linear equation to determine the number of neighbors to be used.

We will compare our method with several alternatives that replace our predictive model, which we refer to as RLAD+K-NN, with a different learning procedure such as LASSO, CART, and OLS+KNN \cite{bertsimas2017personalized}. Both deterministic and randomized prescriptive policies are considered using predictions from these models. To evaluate the performance, we need to determine the effects of the counterfactual treatments. By assessing the predictive power of several models on the non-grouped datasets in terms of their R\textsuperscript{2} and out-of-sample estimation errors (see Appendix), we choose the RLAD+K-NN model that excels in all metrics to impute the outcome for an unobservable treatment $m$, where the number of neighbors should be selected to fit the size of the test set. 

The average improvement in outcomes (the predicted {\em future} outcome under the recommended therapy minus the {\em current} observed outcome) on the test set is shown in Table \ref{tab:results}, where the numbers outside the parentheses are the mean values over 5 repetitions, and the numbers in the parentheses are the corresponding standard deviations. We note that HbA\textsubscript{1c} is measured in percentage while systolic blood pressure in mmHg.
We also list the results from the {\em standard of care}, and the {\em current prescription} which prescribes $m_{\text{f}}(\bx) = m_{\text{c}}(\bx)$ with probability one, i.e., always continuing the current drug regimen.

\begin{table}[hbt]
	\caption{The reduction in HbA\textsubscript{1c}/systolic blood pressure for various models.}
	\label{tab:results}
	\begin{center}
		\begin{tabular}{cccccc}
			\toprule
			&  & \multicolumn{2}{c}{Diabetes} & \multicolumn{2}{c}{Hypertension} \\ \cline{3-6}
			&  & Deterministic & Randomized & Deterministic & Randomized\\
			\midrule
			& LASSO & -0.51 (0.16) & -0.51 (0.16) & -4.71 (1.09) & -4.72 (1.10)\\
			& CART & -0.45 (0.13) & -0.42 (0.14) & -4.84 (0.62) & -4.87 (0.66)\\
			& OLS+K-NN & -0.53 (0.13) & -0.53 (0.13) & -4.33 (0.46) & -4.33 (0.47)\\
			& RLAD+K-NN & -0.56 (0.06) & -0.55 (0.08) & -6.98 (0.86) & -7.22 (0.82) \\  
			\midrule    
			& Current prescription & \multicolumn{2}{c} {-0.22 (0.04)} & \multicolumn{2}{c}{-2.52 (0.19)} \\
			\midrule
			& Standard of care & \multicolumn{2}{c} {-0.22 (0.03)} & \multicolumn{2}{c}{-2.37 (0.11)} \\
			\bottomrule
		\end{tabular}
	\end{center}
\end{table}

Several observations are in order: (i) all models outperform the current prescription and the standard of care; (ii) the RLAD+K-NN model leads to the largest reduction in outcomes with a relatively stable performance; and (iii) the randomized policy achieves a similar performance (slightly better on the hypertension dataset) to the deterministic one. We expect the randomized strategy to win when the effects of several treatments do not differ much, in which case the deterministic algorithm might produce misleading results. The randomized policy could potentially improve the out-of-sample (generalization) performance, as it gives the flexibility of exploring options that are suboptimal on the training set, but might be optimal on the test set. The advantages of the RLAD+K-NN model are more prominent on the hypertension dataset due to the fact that we considered a finer classification of the prescriptions for patients with hypertension, while for diabetic patients, we only distinguish between the oral and injectable prescriptions. 

\section{Conclusions}
We developed a prediction-based randomized prescriptive algorithm that determines the probability of prescribing each treatment based on the predictions from an RLAD+K-NN model which takes into account both the high noise level and the nonlinearity of the data. A deterministic variant was obtained as a special case of the randomized strategy. Theoretical guarantees on the optimality of the prescriptive algorithm and a closed-form expression for the threshold level were provided. The proposed approach was applied to two datasets obtained from the Boston Medical Center: one with diabetic patients and the other for patients with hypertension, providing numerical evidence for the superiority of our algorithm in terms of the improvement in outcomes.

\section*{Acknowledgments}
We thank Henghui Zhu who preprocessed the electronic health records to produce the
datasets we used in this paper to validate the methods. We also thank Dr.\ Theofanie Mela
and Dr.\ Rebecca Mishuris for useful discussions. 


\section*{Appendix}
\paragraph{Proof of Theorem 2.1}
\begin{proof}
	The proof borrows ideas from Theorem 1.5 in \cite{hazan2016introduction}. 
	Define $W_m \triangleq e^{-\xi \hat{y}_m(\bx)}/\sum_{j=1}^M e^{-\xi \hat{y}_j(\bx)}$, and $\phi \triangleq \sum_{m=1}^M e^{-\xi \hat{y}_m(\bx)} e^{-\xi y_m(\bx)}$. Then,
	\begin{equation*}
		\begin{aligned}
			\phi & =  \Big(\sum_{j=1}^M e^{-\xi \hat{y}_j(\bx)}\Big)\sum_{m=1}^M W_m e^{-\xi y_m(\bx)} \\
			       & \le \Big(\sum_{j=1}^M e^{-\xi \hat{y}_j(\bx)}\Big) \sum_{m=1}^M W_m \big(1-\xi y_m(\bx) + \xi^2 y_m^2(\bx)\big) \\
			       & = \Big(\sum_{j=1}^M e^{-\xi \hat{y}_j(\bx)}\Big) \Big( 1 - \xi \sum_{m=1}^M W_m y_m(\bx) + \xi^2 \sum_{m=1}^M W_my_m^2(\bx)\Big) \\
			       & \le \Big(\sum_{j=1}^M e^{-\xi \hat{y}_j(\bx)}\Big) e^{- \xi \sum_{m=1}^M W_my_m(\bx) + \xi^2 \sum_{m=1}^M W_m y_m^2(\bx)},
		\end{aligned}
	\end{equation*}
	where the first inequality uses the fact that for $x\ge 0$, $e^{-x} \le 1-x+x^2$, and the last inequality is due to the fact that $1+x \le e^{x}$. Next let us examine the sum of exponentials:
	\begin{equation*}
		\begin{aligned}
			\sum_{j=1}^M e^{-\xi \hat{y}_j(\bx)} & \le \sum_{j=1}^M \Big( 1 - \xi \hat{y}_j(\bx) + \xi^2\hat{y}^2_j(\bx) \Big) \\
			& = M \Big( 1 - \xi \frac{1}{M}\sum_{j=1}^M \hat{y}_j(\bx) + \xi^2 \frac{1}{M}\sum_{j=1}^M \hat{y}_j^2(\bx)\Big) \\
			& \le M e^{- \xi \frac{1}{M}\sum_{j=1}^M \hat{y}_j(\bx) + \xi^2 \frac{1}{M}\sum_{j=1}^M \hat{y}_j^2(\bx)}.
		\end{aligned}
	\end{equation*}
	On the other hand, for any $k \in [M]$,
	\begin{equation} \label{eq1}
		e^{-\xi \hat{y}_k(\bx) -\xi y_k(\bx)}\le \phi \le M e^{- \xi \frac{1}{M}\sum_{j=1}^M \hat{y}_j(\bx) + \xi^2 \frac{1}{M}\sum_{j=1}^M \hat{y}_j^2(\bx) - \xi \sum_{m=1}^M W_my_m(\bx) + \xi^2 \sum_{m=1}^M W_m y_m^2(\bx)}.
	\end{equation}
	Taking the logarithm on both sides of (\ref{eq1}) and dividing by $\xi$, we obtain
	\begin{equation*}
	\begin{aligned}
	\frac{1}{M} \sum_{m=1}^M \hat{y}_m(\bx) + \sum_{m=1}^M & \frac{e^{-\xi \hat{y}_m(\bx)}}{\sum_j e^{-\xi \hat{y}_j(\bx)}}  y_m(\bx)  \le \hat{y}_k(\bx) + y_k(\bx)  \\
	& + \xi \bigg( \frac{1}{M} \sum_{m=1}^M \hat{y}_m^2(\bx) + \sum_{m=1}^M \frac{e^{-\xi \hat{y}_m(\bx)}}{\sum_j e^{-\xi \hat{y}_j(\bx)}}  y_m^2(\bx)\bigg) + \frac{\log M}{\xi}.
	\end{aligned}
	\end{equation*}
\end{proof}
%
\paragraph{Proof of Theorem 2.2}
\begin{proof}
	We first review the definition of sub-Gaussian random variables.	
	\begin{defi}[Sub-Gaussian random variable]
		A random variable $y \in \mbb{R}$ with mean $\mu_y \triangleq \mbb{E}(y)$ is sub-Gaussian if there exists some positive constant $C$ such that the tail of $y$ satisfies:
		\begin{equation} \label{sgdefi}
		\mbb{P}(|y-\mu_y| \ge t) \le 2\exp(-t^2/(2C^2)), \forall t \ge 0.
		\end{equation}
		The smallest constant $\sqrt{2}C$ satisfying (\ref{sgdefi}) is called the sub-Gaussian norm, or the $\psi_2$-norm of $y$, denoted as $\vertiii{y}_{\psi_2}$.
	\end{defi}
	By the sub-Gaussian assumption we have:
	\begin{equation} \label{tailprobr}
	\begin{aligned}
	\mbb{P}\Big(\sum_k \frac{e^{-\xi \hat{y}_k(\bx)}}{\sum_j e^{-\xi \hat{y}_j(\bx)}}  \hat{y}_k(\bx) > x_{\text{co}} - T(\bx)\Big) & \le \mbb{P}\Bigl(\max\limits_k \hat{y}_k(\bx)> x_{\text{co}} - T(\bx)\Bigr) \\
	& =  \mbb{P}\Big(\bigcup\limits_k \big\{\hat{y}_k(\bx)> x_{\text{co}} - T(\bx)\big\}\Big) \\
	& \le \sum\limits_k \mbb{P} \big(\hat{y}_k(\bx)> x_{\text{co}} - T(\bx)\big) \\
	& \le \sum\limits_k \exp \Bigl(-\frac{\bigl(x_{\text{co}} - T(\bx) - \mu_{\hat{y}_k}(\bx)\bigr)^2}{2 C_k^2(\bx)}\Bigr), \\
	\end{aligned}
	\end{equation} 
	where $\mu_{\hat{y}_k}(\bx) = \mbb{E}[\hat{y}_k(\bx)|\bx]$. Note that the probability in (\ref{tailprobr}) is taken with respect to the measure of the training samples. We essentially want to find the largest threshold $T(\bx)$ such that the probability of the expected improvement being less than $T(\bx)$ is small. Given a small $0 < \bar{\epsilon} < 1$, if
	\begin{equation*}
	\mbb{P}\Bigl(\sum_k \frac{e^{-\xi \hat{y}_k(\bx)}}{\sum_j e^{-\xi \hat{y}_j(\bx)}}  \hat{y}_k(\bx) > x_{\text{co}} - T(\bx)\Bigr) \le \bar{\epsilon},
	\end{equation*}
	then from (\ref{tailprobr}), we get:
	\begin{equation} \label{epsilonbound}
	\sum\limits_k \exp \Bigl(-\frac{\bigl(x_{\text{co}} - T(\bx) - \mu_{\hat{y}_k}(\bx)\bigr)^2}{2 C_k^2(\bx)}\Bigr) \le \bar{\epsilon}.
	\end{equation}
	A relaxation of (\ref{epsilonbound}) implies that:
	\begin{equation*}
	\exp \Bigl(-\frac{\bigl(x_{\text{co}} - T(\bx) - \mu_{\hat{y}_m}(\bx)\bigr)^2}{2 C_m^2(\bx)}\Bigr) \le \frac{\bar{\epsilon}}{M}, \ \forall m,
	\end{equation*}
	which yields that,
	\begin{equation*}
	T(\bx) \le x_{\text{co}} - \mu_{\hat{y}_{m}}(\bx) - \sqrt{-2 C_m^2(\bx)\log (\bar{\epsilon}/M)}, \ \forall m.
	\end{equation*}
	Given that $T(\bx)$ is non-negative, we take:
	\begin{equation*}
	T(\bx) = \max\biggl(0, \ \min_m \Bigl(x_{\text{co}} - \mu_{\hat{y}_{m}}(\bx) - \sqrt{-2 C_m^2(\bx)\log (\bar{\epsilon}/M)}\Bigr)\biggr).
	\end{equation*}
	The parameters $\mu_{\hat{y}_{m}}(\bx)$ and $C_m(\bx)$ for $m = 1, \ldots, M$ can be estimated through Algorithm \ref{ms}.
		
	If using the deterministic policy ($\xi \to \infty$),
	\begin{equation} \label{tailprob}
	\begin{aligned}
	\mbb{P}(\min\limits_m \hat{y}_m(\bx)> x_{\text{co}} - T(\bx)) & = \mbb{P}\Bigl(\bigcap \limits_m \big\{\hat{y}_m(\bx)> x_{\text{co}} - T(\bx)\big\}\Bigr) \\
	& \le  \mbb{P}(\hat{y}_m(\bx)> x_{\text{co}} - T(\bx)) \\
	& \le \exp \Bigl(-\frac{\bigl(x_{\text{co}} - T(\bx) - \mu_{\hat{y}_m}(\bx)\bigr)^2}{2 C_m^2(\bx)}\Bigr), \forall m.\\
	\end{aligned}
	\end{equation} 
    Similarly, to make
	\begin{equation*}
	\mbb{P}\bigl(\min\limits_m \hat{y}_m(\bx)> x_{\text{co}} - T(\bx)\bigr) \le \bar{\epsilon},
	\end{equation*}
	we take
	\begin{equation*}
	T(\bx) = \max\biggl(0, \ \min_m \Bigl(x_{\text{co}} - \mu_{\hat{y}_{m}}(\bx) - \sqrt{-2 C_m^2(\bx)\log \bar{\epsilon}}\Bigr)\biggr).
	\end{equation*}
\end{proof}

\begin{algorithm}
	\caption{Estimating the conditional mean $\mu_{\hat{y}_{m}}(\bx)$ and standard deviation $C_m(\bx)$ of the predicted outcome.} \label{ms}
	\begin{algorithmic}[1]
		\State \textbf{Input:} a feature vector $\bx$; $a_m$: the number of subsamples used to compute $\hat{\bbeta}_m$, $a_m < N_m$; $d_m$: the number of repetitions.		
		\For{$i = 1, \ldots, d_m$}
		\State Randomly pick $a_m$ samples from prescription group $m$, and use them to estimate a robust regression coefficient $\hat{\bbeta}_{m_i}$ through solving (\ref{qcp}).
		\State The future outcome for $\bx$ under prescription $m$ is predicted as $\hat{y}_{m_i}(\bx) = \bx'\hat{\bbeta}_{m_i}$.
		\EndFor
		\State Estimate the conditional mean of $\hat{y}_m(\bx)$ as: $$\mu_{\hat{y}_{m}}(\bx) = \frac{1}{d_m} \sum_{i=1}^{d_m}\hat{y}_{m_i}(\bx),$$
		and the conditional standard deviation as:
		\begin{equation*}
		C_m(\bx) = \sqrt{\frac{1}{d_m - 1} \sum_{i=1}^{d_m} \Bigl(\hat{y}_{m_i}(\bx) - \mu_{\hat{y}_{m}}(\bx)\Bigr)^2}.
		\end{equation*}
	\end{algorithmic}
\end{algorithm}

\paragraph{Predictive Performance of Various Models}
We use four metrics to evaluate the predictive power of various models on the test set:
\begin{itemize}
	\item The R-square:
	\begin{equation*}
	\text{R\textsuperscript{2}} (\by, \hat{\by}) = 1-\dfrac{\sum_{i=1}^{N_t} (y_{i}-\hat{y}_{i})^2}{\sum_{i=1}^{N_t} (y_{i}-\bar{y})^2},
	\end{equation*}
	where $\by = (y_{1},\,\ldots,\, y_{N_t})$ and $\hat{\by} = (\hat{y}_{1},\,\ldots,\, \hat{y}_{N_t})$ are the vectors of the true (observed) and predicted outcomes, respectively, with $N_t$ the size of the test set, and $\bar{y} = (1/N_t) \sum_{i=1}^{N_t} y_i$. 
	\item The {\em Mean Squared Error (MSE)}: 
	\begin{equation*}
	\text{MSE} (\by, \hat{\by}) = \frac{1}{N_t} \sum_{i=1}^{N_t} (y_{i}-\hat{y}_{i})^2.
	\end{equation*}
	\item The {\em Mean Absolute Error (MeanAE)} that is more robust to large deviations than the MSE in that the absolute value function increases more slowly than the square function over large (absolute) values of the argument.
	\begin{equation*}
	\text{MeanAE} (\by, \hat{\by}) = \frac{1}{N_t} \sum_{i=1}^{N_t} |y_{i}-\hat{y}_{i}|.
	\end{equation*}
	\item The MedianAE which can be viewed as a robust measure of the MeanAE, computing the median of the absolute deviations:
	\begin{equation*}
	\text{MedianAE} (\by, \hat{\by}) = \text{Median} \left( |y_{i}-\hat{y}_{i}| , i = 1, \ldots, N_t\right).
	\end{equation*}	
\end{itemize}

The out-of-sample performance metrics of the various models on the two datasets are shown in Tables \ref{tab:diabetes_preformances} and \ref{tab:hypertension}, where the numbers in the parentheses show the improvement of RLAD+K-NN compared against other methods. Huber refers to the robust regression method proposed in \cite{huber1964robust, huber1973robust}, and CART refers to the {\em Classification And Regression Trees}. Huber/OLS/LASSO + K-NN means fitting a K-NN regression model with a Huber/OLS/LASSO-weighted distance metric. We note that in order to produce well-defined and meaningful predictive performance metrics, the datasets used to generate Tables \ref{tab:diabetes_preformances} and \ref{tab:hypertension} did not group the patients by their prescriptions. A universal model was fit to all patients using the prescription as one of the predictors. Nevertheless, it would still be considered as a fair comparison as all models were evaluated on the same dataset. The results provide supporting evidence for the validity of our RLAD+K-NN model that outperforms all others in all metrics, and is thus used for predicting the outcomes of counterfactual treatments.

\begin{table}[htbp]
	{\centering 
		\caption{Performance of different models for predicting future HbA\textsubscript{1c} for diabetic patients.} 
		\label{tab:diabetes_preformances} 
		\begin{tabular}{cccccc}
			\toprule
			Methods & R\textsuperscript{2} & MSE & MeanAE & MedianAE\\
			\midrule
			OLS  & 0.52 (2\%) & 1.36 (2\%) & 0.81 (4\%) & 0.55 (11\%)\\
			LASSO  & 0.52 (2\%) & 1.37 (2\%) & 0.80 (3\%) & 0.54 (9\%)\\
			Huber  & 0.36 (47\%) & 1.81 (26\%) & 0.96 (19\%) & 0.70 (30\%)\\
			RLAD  & 0.50 (4\%) & 1.40 (4\%) & 0.78 (1\%) & 0.50 (1\%)\\
			K-NN  & 0.25 (109\%) & 2.11 (37\%) & 1.07 (27\%) & 0.81 (39\%)\\
        OLS+K-NN  & 0.52 (0\%) & 1.34 (0\%) & 0.79 (1\%) & 0.51 (3\%)\\
        LASSO+K-NN  & 0.52 (1\%) & 1.36 (1\%) & 0.79 (1\%) & 0.50 (1\%)\\
			Huber+K-NN  & 0.51 (3\%) & 1.38 (3\%) & 0.81 (3\%) & 0.53 (7\%)\\
			RLAD+K-NN  & 0.52 (N/A) & 1.34 (N/A) & 0.78 (N/A) & 0.49 (N/A)\\
			CART  & 0.49 (7\%) & 1.43 (7\%) & 0.81 (3\%) & 0.50 (2\%)\\
			\bottomrule
		\end{tabular}\\
	}
\end{table}

\begin{table}[htbp]
	{\centering 
		\caption{Performance of different models for predicting future systolic blood pressure for hypertension patients. }
		\label{tab:hypertension} 
		\begin{tabular}{ccccc}
			\toprule
			Methods & R\textsuperscript{2} & MSE & MeanAE & MedianAE\\
			\midrule
			OLS  & 0.31 (14\%) & 170.80 (6\%) & 10.09 (7\%) & 8.15 (9\%)\\
			LASSO  & 0.31 (14\%) & 170.83 (6\%) & 10.08 (7\%) & 8.22 (10\%)\\
			Huber  & 0.22 (62\%) & 193.54 (17\%) & 10.70 (12\%) & 8.61 (14\%)\\
			RLAD  & 0.30 (18\%) & 173.32 (8\%) & 10.11 (7\%) & 8.28 (11\%)\\
			K-NN  & 0.33 (10\%) & 167.41 (5\%) & 9.62 (2\%) & 7.50 (2\%)\\
			     OLS+K-NN  & 0.35 (1\%) & 160.22 (0\%) & 9.42 (0\%) & 7.49 (1\%)\\
			     LASSO+K-NN  & 0.32 (12\%) & 169.50 (6\%) & 9.74 (3\%) & 7.73 (5\%)\\
			Huber+K-NN  & 0.32 (10\%) & 167.92 (5\%) & 9.71 (3\%) & 7.84 (6\%)\\
			RLAD+K-NN  & 0.36 (N/A) & 159.74 (N/A) & 9.42 (N/A) & 7.38 (N/A)\\
			CART  & 0.25 (43\%) & 186.23 (14\%) & 10.34 (9\%) & 8.22 (10\%)\\
			\bottomrule
		\end{tabular}\\
	}
\end{table}

\begin{table}[htbp]
	\centering 
	\caption{Feature importance from the RLAD model for the diabetes dataset.} 
	\label{tab:importance_diabetes}
	\begin{tabular}{cc}
		\toprule
		Features & Regression coefficients\\
		\midrule
		lab test: HbA\textsubscript{1c}  & 1.02\\
		lab test: blood glucose  & 0.27\\
		measurement: systolic blood pressure  & 0.07\\
		prescription: injectable  & 0.05\\
		lab test: hematocrit  & 0.04\\
		lab test: hemoglobin  & -0.03\\
		lab test: sodium  & 0.03\\
		lab test: platelet count  & 0.03\\
		diagnosis: retinal disorders  & 0.02\\
		lab test: leukocyte count  & 0.02\\
		diagnosis: inflammatory and toxic neuropathy  & 0.02\\
		lab test: mean corpuscular volume  & -0.02\\
		lab test: calcium  & 0.01\\
		lab test: potassium  & 0.01\\
		diagnosis: disorders of lipoid metabolism  & 0.01\\
		diagnosis: other disorders of soft tissues  & 0.01\\
		race: Caucasian  & -0.01\\
		diagnosis: obesity  & -0.01\\
		age  & -0.01\\
		diagnosis: essential hypertension  & 0.01\\
		\bottomrule
	\end{tabular}
	\label{tab:diabetes_features} 
\end{table}

\begin{table}[htbp]
	\centering 
	\caption{Feature importance from the RLAD model for the hypertension dataset.} 
	\label{tab:importance_hypertension} 
	\begin{tabular}{cc}
		\toprule
		Features & Regression coefficients\\
		\midrule
		measurement: systolic blood pressure  & 7.62\\
		age  & 1.87\\
		lab test: sodium  & 1.29\\
		lab test: hemoglobin  & 1.26\\
		prescription: calcium channel blockers  & 0.98\\
		lab test: blood glucose  & 0.93\\
		lab test: hematocrit  & -0.82\\
		sex: female  & 0.76\\
		lab:mean corpuscular volume  & -0.61\\
		diagnosis: asthma  & -0.61\\
		prescription: ARB  & 0.57\\
		diagnosis: cataract  & 0.57\\
		diagnosis: chronic ischemic heart disease  & -0.56\\
		lab test: potassium  & 0.55\\
		diagnosis: heart failure  & -0.53\\
		prescription: diuretics  & 0.53\\
		diagnosis: cardiac dysrhythmias  & -0.51\\
		diagnosis obesity  & 0.46\\
		race: Caucasian  & -0.46\\
		diagnosis: disorders of fluid electrolyte and acid-base balance  & 0.45\\
		\bottomrule
	\end{tabular}
\end{table}

\paragraph{Oral prescriptions for diabetes.} Acarbose, acetohexamide, chlorpropamide, glimepiride, glipizide, glyburide, hydrochloride, metformin, miglitol, nateglinide, pioglitazone, repaglinide, rosiglitazone, sitagliptin, tolazamide, tolbutamide, and troglitazone.

\end{document}